\documentclass[letterpaper, 10 pt, conference]{ieeeconf}  
\IEEEoverridecommandlockouts                              
\usepackage{epsfig} 
\usepackage{times} 
\usepackage{hyperref}
\usepackage{mathtools,tikz}
\usepackage{tikzscale}
\usepackage{balance}
\usepackage{xcolor}
\usepackage{pgfplots,graphicx}
\usepackage{amsmath} 
\usepackage{amssymb}  
\usepackage{mathrsfs}
\usepackage{comment} 
\usepackage{algorithm,algorithmic}

 \DeclareMathOperator*{\argmin}{argmin}

\newtheorem{theorem}{\protect\theoremname}
\newtheorem{defn}{\protect\definitionname}
\newtheorem{proposition}{\protect\propositionname}

\providecommand{\definitionname}{\textbf{Definition}}
\providecommand{\propositionname}{\textbf{Proposition}}
\providecommand{\remarkname}{\textbf{Remark}}
\providecommand{\theoremname}{\textbf{Theorem}}
\providecommand{\lemmaname}{Lemma}
\providecommand{\assumname}{\textbf{Assumption}}
\providecommand{\probname}{\textbf{Problem}}

\title{\Large \bf
 Risk-Averse Planning via CVaR Barrier Functions:\\ Application to Bipedal Robot Locomotion
\author{Mohamadreza Ahmadi, Xiaobin Xiong, and Aaron D. Ames
\thanks{The authors are with the Department of Mechanical and Civil Engineering, California Institute of Technology, Pasadena, CA 91125. email: {\tt(\{mrahmadi,xxiong,ames\}@caltech.edu)}.}
 }}
\begin{document}
\maketitle
\thispagestyle{empty}
\pagestyle{empty}

\begin{abstract}
Enforcing safety  in the presence of stochastic uncertainty is a challenging problem. Traditionally, researchers have proposed safety in the statistical mean as a safety measure in this case. However, ensuring safety in the statistical mean is only reasonable if system's safe behavior in the large number of runs is of interest, which precludes the use of mean safety in practical scenarios. In this paper, we propose a risk sensitive notion of safety called conditional-value-at-risk (CVaR) safety, which is concerned with safe performance in the worst case realizations. We introduce CVaR barrier functions as a tool to enforce CVaR-safety and propose conditions for their Boolean compositions. Given a legacy controller, we show that we can design a \textit{minimally interfering} CVaR-safe controller via solving difference convex programs. {We elucidate the proposed method by applying it to a bipedal robot locomotion case study.}
\end{abstract}

\IEEEpeerreviewmaketitle

\section{INTRODUCTION}

With the rise of autonomous systems being deployed in real-world settings, the associated risk that stems from uncertain and unforeseen circumstances is correspondingly on the rise. For instance, there are several inherent sources of uncertainty in robotics systems, such as modeling uncertainty, sensor range and resolution limitations,  dynamic and uncertain environments, noise and wear-and-tear in robot actuation~\cite{thrun2005probabilistic}, that lead to higher risk during deployment.


Mathematically speaking, risk can be quantified in numerous ways, such as chance constraints~\cite{ono2015chance,wang2020non}. However, applications in autonomy and robotics require more ``nuanced assessments of risk''~\cite{majumdar2020should}. Artzner \textit{et. al.}~\cite{artzner1999coherent} characterized a set of natural properties that are desirable for a risk measure, called a coherent risk measure, and  have  obtained widespread
acceptance in finance and operations research, among other fields. An important example of a coherent risk measure is the conditional value-at-risk (CVaR) that has received significant attention in decision making problems, such as Markov decision processes (MDPs)~\cite{chow2015risk,chow2014algorithms,bauerle2011markov}.  For stochastic discrete-time dynamical systems, a model predictive control technique with coherent risk objectives was proposed in~\cite{singh2018framework}, wherein the authors also proposed Lyapunov conditions for risk-sensitive exponential stability. Moreover, a method based on stochastic reachability analysis was proposed in~\cite{chapman2019risk} to estimate a CVaR-safe set of initial conditions via the solution to an MDP.

In this work, we use a special class of barrier functions as a tool for enforcing risk-sensitive safety. Control barrier functions have been proposed in~\cite{ames2016control} and have been used for designing safe controllers (in
the absence of a legacy controller,\textit{ i.e.}, a desired controller that may be unsafe) and safety filters (in
the presence of a legacy controller) for continuous-time dynamical systems, such as bipedal robots~\cite{nguyen20163d}, with guaranteed robustness~\cite{xu2015robustness,kolathaya2018input}. For discrete-time systems, discrete-time barrier functions have been formulated in~\cite{ahmadi2019safe,agrawal2017discrete} and applied to the multi-robot coordination problem~\cite{ahmadi2020barrier}. Recently, for a class of stochastic (Ito) differential equations, safety in probability and statistical mean were also studied in~\cite{clark2019control,santoyo2019barrier,luo2019multi}.


\begin{figure}[t] \centering{
\includegraphics[scale=.27]{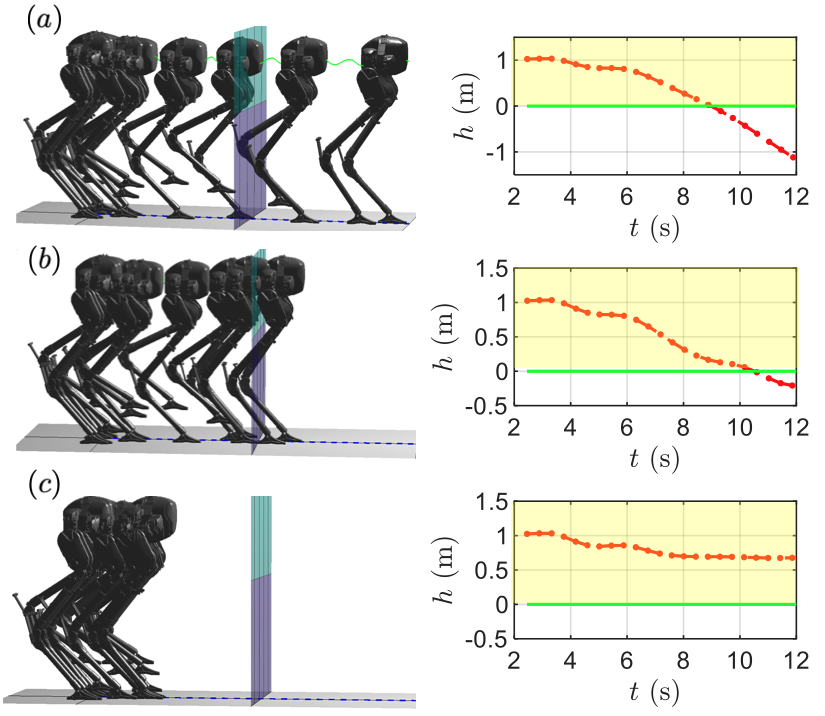}
\vspace{-.3cm}
\caption{Risk-averse obstacle avoidance using CVaR barrier functions (robot behavior and barrier function evolution). The shaded yellow area denotes safe regions. (a) safety violation with no barrier function; (b) safety violation with risk-neutral barrier function; (c) safe behavior with CVaR barrier function. Plots on the right side show the values of the barrier functions.}\label{fig:fixedwall}
}
 \end{figure}


In this paper, we go beyond the conventional notions of safety in probability and statistical mean for discrete-time systems subject to stochastic uncertainty. To this end, we define safety in the risk-sensitive CVaR sense, which is concerned with safety in the worst possible scenarios. We then propose CVaR barrier functions as a tool to enforce CVaR-safety and formulate conditions for their Boolean compositions. We propose a computational method based on difference convex programs (DCPs) to synthesize CVaR-safe controllers for stochastic linear discrete-time systems. These CVaR-safe controllers are designed such that they minimally interfere with a given  legacy controller. {We show the efficacy of our proposed method on collision avoidance scenarios involving a bipedal robot subject to modeling uncertainty~(see Figure~\ref{fig:fixedwall}).}

The rest of the paper is organized as follows. In the next section, we introduce CVaR-safety and formulate CVaR barrier functions as a tool to synthesize risk-averse safe controllers. In Section~III, we discuss our robot bipedal locomotion case study and present the obtained results. Finally, in Section~V, we conclude the paper and give directions for future research.

\textbf{Notation: } We denote by $\mathbb{R}^n$ the $n$-dimensional Euclidean space and $\mathbb{N}_{\ge0}$ the set of non-negative integers.  For a finite set $\mathcal{A}$, we denote by $|\mathcal{A}|$ the number of elements of $\mathcal{A}$. For  a probability space $(\mathcal{X}, \mathcal{F}, \mathbb{P})$ and a constant $p \in [1,\infty)$, $\mathcal{L}_p(\mathcal{X}, \mathcal{F}, \mathbb{P})$ denotes the vector space of real valued random variables $X$ for which $\mathbb{E}|X|^p < \infty$. The Boolean operators are denoted by $\neg$ (negation), $\lor$ (conjunction), and $\land$ (disjunction). For a risk measure $\rho$, we denote $\rho^t$ to show the function composition of $\rho$ with itself $t$ times.



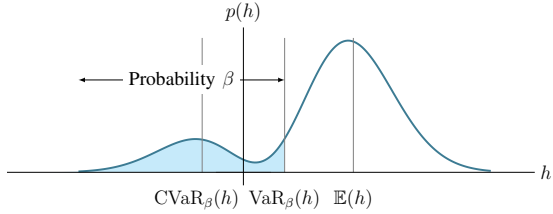
\begin{figure} 
\centering{
    \resizebox{0.42\textwidth}{!}{
\begin{tikzpicture}
\tikzstyle{every node}=[font=\large]
\pgfmathdeclarefunction{gauss}{2}{%
  \pgfmathparse{1/(#2*sqrt(2*pi))*((x-.5)^2+.5)*exp(-((x-#1)^2)/(2*#2^2))}%
}

\pgfmathdeclarefunction{gauss2}{3}{%
\pgfmathparse{1/(#2*sqrt(2*pi))*((#1-.5)^2+.5)*exp(-((#1-#1)^2)/(2*#2^2))}
}

\begin{axis}[
  no markers, domain=-6:9, range=-2:8, samples=200,
  axis lines*=center, xlabel=$h$, ylabel=$p(h)$,
  every axis y label/.style={at=(current axis.above origin),anchor=south},
  every axis x label/.style={at=(current axis.right of origin),anchor=west},
  height=5cm, width=13cm,
  xtick={0}, ytick=\empty,
  enlargelimits=true, clip=false, axis on top,
  grid = major
  ]
  \addplot [fill=cyan!20, draw=none, domain=-7:1.5] {gauss(1.5,2)} \closedcycle;
  \addplot [very thick,cyan!50!black] {gauss(1.5,2)};
 
 \pgfmathsetmacro\valueA{gauss2(1.5,1.5,2)}
 \draw [gray] (axis cs:1.5,0) -- (axis cs:1.5,\valueB);
  \pgfmathsetmacro\valueB{gauss2(-1.5,1.5,2)}
  \draw [gray] (axis cs:-1.5,0) -- (axis cs:-1.5,\valueB);
    \draw [gray] (axis cs:4,0) -- (axis cs:4,\valueB);
 
 
 \draw [gray] (axis cs:1,0)--(axis cs:-1,0);
\node[below] at (axis cs:1.5, -0.1)  {$\mathrm{VaR}_\beta(h)$}; 
\node[below] at (axis cs:4, -0.1)  {$\mathbb{E}(h)$}; 
\node[below] at (axis cs:-1.7, -0.1)  {$\mathrm{CVaR}_\beta(h)$}; 
\draw [yshift=2cm, latex-latex](axis cs:-6,0) -- node [fill=white] {Probability $\beta$} (axis cs:1.5,0);
\end{axis}
\end{tikzpicture}
}
}
\caption{Comparison of the mean, VaR, and CVaR for a given confidence $\beta \in (0,1)$. The axes denote the values of the stochastic variable $h$ and its  pdf $p(h)$. The shaded area denotes the $\%\beta$ of the area under $p(h)$. If $h\ge0$ represents a safe behavior, using $\mathbb{E}(h)$ as a performance measure is misleading (note that $\mathbb{E}(h)$ is positive).  VaR gives the value of $h$ at the $\beta$-tail of the distribution. But, it ignores the values of $h$ with probability below $\beta$.  CVaR is the average of the values of VaR with probability less than $\beta$ (average of the worst-case values of $h$ in the $\beta$ tail of the distribution). Note that $\mathrm{CVaR}_\beta$ is negative indicating unsafe behavior.} \label{fig:varvscvar}
\end{figure}

\section{CVaR Barrier Functions for\\ Risk-Averse Planning}
\label{overview}

In this section, we formulate the risk-averse safety problem and propose a solution based on a special class of barrier functions. We begin by defining our risk measure of interest called CVaR. 

\subsection{Conditional Value-at-Risk}

Let $(\Omega, \mathcal{F}, \mathbb{P})$ be a probability space, $\mathcal{H} = \mathcal{L}_p(\Omega, \mathcal{F}, \mathbb{P})$, $p \in [0,\infty)$, and let $h \in \mathcal{H}$ be a stochastic variable for which higher values are of interest (for example, greater values of $h$ indicate safer performance). For a given confidence level $\beta \in (0,1)$, value-at-risk ($\mathrm{VaR}_{\beta}$) denotes the $\beta$-quantile value of a stochastic  variable $h \in \mathcal{H}$ described as
$
\mathrm{VaR}_{\beta}(h) = \sup_{\zeta \in \mathbb{R}} \{ \zeta \mid \mathbb{P}(h \le \zeta ) \le \beta \}.
$ 

Unfortunately, working with VaR  for non-normal stochastic variables is numerically unstable, optimizing models involving  VaR are intractable in
high dimensions, and VaR ignores the values of $h$ with probability less than $\beta$~\cite{rockafellar2000optimization}. 

In contrast, CVaR overcomes the shortcomings of VaR. CVaR with confidence level $\beta \in (0,1)$ denoted $\mathrm{CVaR}_{\beta}$ measures the expected loss in the $\beta$-tail given that the particular threshold $\mathrm{VaR}_{\beta}$ has been crossed,\textit{ i.e.}, $\mathrm{CVaR}_{\beta} (h) =  \mathbb{E}\left[ h \mid h \le \mathrm{VaR}_{\beta}(h)  \right]$. An optimization formulation for CVaR was proposed in~\cite{rockafellar2000optimization} that we use in this paper. That is, $\mathrm{CVaR}_{\beta}$ is given by 
\begin{align}
    \mathrm{CVaR}_{\beta}(h):=&\inf_{\zeta \in \mathbb{R}}\mathbb{E}\left[\zeta + \frac{(h-\zeta)_{+}}{\beta}\right], \label{eq:cvardual}
\end{align}
 where $(\cdot)_{+}=\max\{\cdot, 0\}$. A value of $\beta \to 1$ corresponds to a risk-neutral case,\textit{ i.e.},  $\mathrm{CVaR_1}(h)=\mathbb{E}(h)$; whereas, a value of $\beta \to 0$ is rather a risk-averse case,\textit{ i.e.}, $\mathrm{CVaR_0}(h)=\mathrm{VaR}_0(h)$~\cite{rockafellar2002conditional}. Figure~\ref{fig:varvscvar} illustrates these notions for an example $h$ variable with distribution $p(h)$.
 


Unlike VaR,  CVaR  is a coherent risk measure~\cite{shapiro2014lectures}, which satisfies the following properties. 

\vspace{0.2cm}
\begin{defn}[Coherent Risk Measure]\label{defi:coherent}\textit{
We call a risk measures $\rho: \mathcal{H} \to \mathbb{R}$ a \emph{coherent risk measure}, if it satisfies the following conditions
\begin{itemize}
    \item \textbf{Convexity:} $\rho(\lambda h + (1-\lambda)h') \le \lambda \rho(h)+(1-\lambda)\rho(h')$, for all $\lambda \in (0,1)$ and for all $h,h' \in \mathcal{H}$;
    \item \textbf{Monotonicity:} If $h\le h'$ then $\rho(h) \le \rho(h')$ for all $h,h' \in \mathcal{H}$;
    \item \textbf{Translational Invariance:} $\rho(h+c)=\rho(h)+c$ for all $h \in \mathcal{H}$ and $c \in \mathbb{R}$;
    \item \textbf{Positive Homogeneity:} $\rho(\beta h)= \beta \rho(h)$ for all $h \in \mathcal{H}$ and $\beta \ge 0$.
\end{itemize}
}
\end{defn}

In fact, we use the nice mathematical properties of CVaR given in Definition 1 in the proofs of our main results in Section~\ref{sec:main}.

\subsection{CVaR-Safety}

We consider discrete-time stochastic systems  given by
\begin{equation}\label{eq:dynamics}
    x^{t+1} = f(x^t,u^t,w^t), \quad x^0=x_0,
\end{equation}
where $t \in \mathbb{N}_{\ge 0}$ denotes the time index, $x \in \mathcal{X} \subset \mathbb{R}^n$ is the state, $u \in \mathcal{U} \subset \mathbb{R}^m$ is the control input,  $w \in \mathcal{W}$ is the stochastic uncertainty/disturbance, and the (possibly nonlinear) function $f:  \mathbb{R}^n \times \mathcal{U} \times \mathcal{W} \to \mathbb{R}^n$. We assume that the initial condition $x_0$ is deterministic and that $|\mathcal{W}|$ is finite, \textit{i.e.,} $\mathcal{W} = \{w_1, \ldots, w_{|\mathcal{W}|}\}$. At every time-step $t$,
for a state-control pair $(x^t, u^t)$, the process disturbance $w^t$ is
drawn from set $\mathcal{W}$ according to the probability density function $p(w) = [p(w_1),\ldots,p(w_{|\mathcal{W}|})]^T$, where $p(w_i):=\mathbb{P}(w^t=w_i)$, $i=1,2,\ldots,|\mathcal{W}|$. Note that the
probability mass function for the process disturbance is time-invariant, and that the process disturbance is independent of
the process history and of the state-control pair $(x^t, u^t)$.

We are interested in studying the properties of the solutions to~\eqref{eq:dynamics} with respect to the compact set $\mathcal{S}$ described as \begin{subequations}\label{eq:safeset}
\begin{eqnarray}
\mathcal{S} :=\{ x \in \mathcal{X} \mid h(x) \ge 0 \}, \\
\mathrm{Int}(\mathcal{S}) :=\{ x \in \mathcal{X} \mid h(x) > 0 \}, \\
\partial \mathcal{S} :=\{ x \in \mathcal{X} \mid h(x) = 0 \},
\end{eqnarray}
\end{subequations}
where $h:\mathcal{X} \to \mathbb{R}$ is a continuous function. For instance, $\mathcal{S}$ can represent robot constraints, e.g. joint limits, safe exploration regions, and etc.

In the presence of stochastic uncertainty $w$, assuring almost sure (with probability one) invariance or safety may not be feasible. Moreover, enforcing safety in expectation is only meaningful if the law of large numbers  can  be  invoked  and  we  are  interested  in  the  long term  performance,  independent  of  the  realization  fluctuations. In this work, instead, we propose safety in a dynamic coherent risk measure, namely, CVaR sense, with conditional expectation (risk-neutral case) as an special case $\beta \to 1$. 
\vspace{0.1cm}
\begin{defn} [CVaR-safety]
\textit{Given a safe set $\mathcal{S}$ as given in~\eqref{eq:safeset} and a  confidence level $\beta \in(0,1)$, we call the solutions to~\eqref{eq:dynamics} starting at $x_0 \in \mathcal{S}$  CVaR-safe  if and only if 
\begin{equation}\label{eq:risksafety}
    \mathrm{CVaR}_{\beta}^t \left( h(x^t) \right) \ge 0, \quad \forall t\ge 0.
\end{equation}
}
\end{defn}
\vspace{0.2cm}

Note that $\mathrm{CVaR}_{\beta}^t$ is a dynamic time-consistent risk measure~\cite[Definition 3]{ruszczynski2010risk},\textit{ i.e.}, if for some two realizations $w$ and $w'$, $h_w(x^\theta) \ge h_{w'}(x^\theta)$ at some future time $\theta$, and $h_w(x^t) = h_{w'}(x^t)$ for time $t \in (\tau, \theta)$, then $h_w(x^t) 	\nless h_{w'}(x^t)$ for $t<\tau$. The time consistency property ensures that contradictory evaluations of safety risk at different points in time does not happen. In other words, if one realization of $w$ incurs higher safety risk at some point in time, then it is a riskier in terms of safety at  any prior point in time. 


\subsection{CVaR Barrier Functions}\label{sec:main}

In order to check and enforce CVaR-safety, we define \emph{CVaR barrier functions}.
\vspace{0.1cm}
\begin{defn}[CVaR Barrier Function]\label{def:riskbf}
\textit{
For~the discrete-time system~\eqref{eq:dynamics} and a confidence level $\beta \in (0,1)$, the continuous function $h : \mathbb{R}^n \to \mathbb{R}$ is a CVaR barrier
function for the set $\mathcal{S}$ as defined in~\eqref{eq:safeset}, if there exists a constant $\alpha \in (0,1)$   such that
\begin{equation}\label{eq:BFinequality}
   \mathrm{CVaR}_{\beta}( h(x^{t+1})) \ge  \alpha h(x^{t}),\quad \forall x^t \in \mathcal{X}.
    \end{equation}
    }
    \end{defn}
\vspace{0.1cm}


In the next result, we demonstrate that the existence of a CVaR barrier function indeed implies CVaR-safety.

\vspace{0.1cm}
\begin{theorem}\label{thm:riskbf}
\textit{
Consider the discrete-time system~\eqref{eq:dynamics} and the set $\mathcal{S}$ as described in~\eqref{eq:safeset}. Let $\beta \in (0,1)$ be a given confidence level. Then, $\mathcal{S}$ is CVaR-safe, if there exists an CVaR barrier function as defined in Definition~\ref{def:riskbf}.
}
\end{theorem}
\vspace{0.1cm}
\begin{proof}
The proof is carried out by induction and using the properties of CVaR as a coherent risk measure as outlined in Definition~\ref{defi:coherent}. If~\eqref{eq:BFinequality} holds, for $t=0$, we have 
\begin{equation}\label{ewqwq}
\mathrm{CVaR}_{\beta}(h(x^1))\ge \alpha  h(x_0).
\end{equation}
Similarly, for $t=1$, we have 
\begin{equation}\label{sdsdd}
\mathrm{CVaR}_{\beta}(h(x^2))\ge \alpha  h(x_1).
\end{equation}
Since $\mathrm{CVaR}_{\beta}$ is monotone (because it is a coherent risk measure), composing both sides of~\eqref{sdsdd} with $\mathrm{CVaR}_{\beta}$ does not change the inequality and we obtain
\begin{equation} \label{eq:bnndds}
\mathrm{CVaR}_{\beta}^2 (h(x^2))\ge  \mathrm{CVaR}_{\beta} (\alpha( h(x^1))).
\end{equation}
Since $\alpha \in (0,1)$, from positive homogeneity property of CVaR, we obtain $\mathrm{CVaR}_{\beta} (\alpha( h(x^1))) = \alpha \mathrm{CVaR}_{\beta} ( h(x^1))$. Thus,~\eqref{eq:bnndds} simplifies to 
$
\mathrm{CVaR}_{\beta}^2 (h(x^2))\ge  \alpha \mathrm{CVaR}_{\beta} ( h(x^1)).
$ 


Then, using inequality~\eqref{ewqwq}, we have 
\begin{equation*}\label{sdsdd2}
\mathrm{CVaR}^2_{\beta} \ge  \alpha  \mathrm{CVaR}_{\beta} (h(x^1)) \ge \alpha^2 h(x_0).
\end{equation*}
Therefore, by induction, at time $t$, we can show that
\begin{equation} \label{eq:expdecay}
\mathrm{CVaR}_{\beta}^t (h(x^t)) \ge \alpha^t  h(x_0).
\end{equation}
If $x_0 \in \mathcal{S}$, from the definition of the set $\mathcal{S}$, we have $h(x_0)\ge0$. Since $\alpha \in (0,1)$, then we can infer that~\eqref{eq:risksafety} holds. Thus, the system is $\mathrm{CVaR}_{\beta}$-safe.
\end{proof}
\vspace{0.2cm}


In many practical  path planning scenarios, we encounter multiple obstacles and safe sets composed of Boolean compositions of several barrier functions~\cite{glotfelter2017nonsmooth,ahmadi2020barrier,ahmadi2020safe}. Next, we propose conditions for checking Boolean compositions of CVaR barrier functions. 
\vspace{.2cm}
\begin{proposition} \label{prop:booleanBFs}
\textit{
Let $\mathcal{S}_i = \{ x \in \mathbb{R}^n \mid h_i(x)\ge 0\}$, $i=1,\ldots,k$ denote a family of safe sets with the boundaries and interior defined analogous to $\mathcal{S}$ in~\eqref{eq:safeset}. Consider the discrete-time system~\eqref{eq:dynamics}. If there exist a $\alpha \in (0,1)$ such that
\begin{equation}\label{eq:disjBF1}
 \textstyle \mathrm{CVaR}_{\beta}\left(\min_{i=1,\ldots,k} h_i(x^{t+1})\right)  \ge  \alpha\min_{i=1,\ldots,k}  h_i(x^{t})
\end{equation}
then the set $\{ x \in \mathbb{R}^n \mid  \land_{i =1,\ldots,k} \left(h_i(x) \ge 0\right)\}$ is CVaR-safe. Similarly, if there exist a $\alpha \in (0,1)$  such that
\begin{equation}\label{eq:disjBF2}
 \textstyle \mathrm{CVaR}_{\beta}\left(\max_{i=1,\ldots,k} h_i(x^{t+1})\right) \ge  \alpha\max_{i=1,\ldots,k}  h_i(x^{t})
\end{equation}
then the set $\{ x  \in \mathbb{R}^n \mid  \lor_{i =1,\ldots,k} \left(h_i(x) \ge 0\right)\}$ is CVaR-safe.
}
\end{proposition}
\vspace{.3cm}
\begin{proof}
 If~\eqref{eq:disjBF1} holds from the proof of Theorem 1, we can infer that 
$$
 \textstyle \mathrm{CVaR}_{\beta}^t \left(\min_{i=1,\ldots,k} h_i(x^{t})\right) \ge \alpha^t\min_{i=1,\ldots,k} h_i(x^0).
$$
That is, if $x^0 \in \{ x \in \mathbb{R}^n \mid \min_{i=1,\ldots,k} h_i(x) \ge 0\}$, then $\mathrm{CVaR}_{\beta}^t \left(\min_{i=1,\ldots,k} h_i(x^{t})\right)\ge 0$ for all $t\in \mathbb{N}_{\ge 0}$. Let $h_{i^*}(x^t)$ be the smallest among $h_i(x^t)$, $i=1,2,...,k$,\textit{ i.e.}, it satisfies 
$
h_j(x^t) \ge \cdots \ge h_{i^*}(x^t),~~\forall j \neq i^*.
$ 
Because CVaR is monotone (see Definition 1), the latter inequality implies 
$
\mathrm{CVaR}_{\beta}^t (h_j(x^t)) \ge \cdots \ge \mathrm{CVaR}_{\beta}^t(h_{i^*}(x^t)),~~\forall j \neq i^*.
$ 
Since $\mathrm{CVaR}_{\beta}^t \left(\min_{i=1,\ldots,k} h_i(x^{t})\right)= \mathrm{CVaR}_{\beta}^t \left( h_{i^*}(x^{t})\right)\ge 0$ for all $t\in \mathbb{N}_{\ge 0}$, we have
$$
\mathrm{CVaR}_{\beta}^t (h_j(x^t)) \ge \cdots \ge \mathrm{CVaR}_{\beta}^t(h_{i^*}(x^t)) \ge 0,~~j \neq i.
$$
Thus, $\mathrm{CVaR}_{\beta}^t(h_{i}(x)) \ge 0$ for all $i \in \{1,\ldots,k\}$.


Similarly, if~\eqref{eq:disjBF2} holds, we can infer that
$$
\textstyle \mathrm{CVaR}_{\beta}^t \left(\max_{i=1,\ldots,k} h_i(x^{t})\right) \ge \alpha^t\max_{i=1,\ldots,k} h_i(x^0).
$$
Hence, using similar arguments as the proof of the conjunction case, $\mathrm{CVaR}_{\beta}^t \left(\max_{i=1,\ldots,k} h_i(x^{t})\right) \ge 0$ for all $t\in \mathbb{N}_{\ge 0}$. That is, there exists at least an $i \in \{1,\ldots,k\}$ for which $\mathrm{CVaR}_{\beta}^t \left(\max_{i=1,\ldots,k} h_i(x^{t})\right) \ge 0$.
\end{proof}
\vspace{.3cm}

The negation operator is trivial and can be shown by checking if $-h$ satisfies inequality~\eqref{eq:BFinequality}.

In the next section, we demonstrate how a sequence $\{ u^t\}_{t>0}$ can be designed such that system~\eqref{eq:dynamics} becomes CVaR-safe based on optimization techniques.

\subsection{CVaR-Safe Controller Synthesis}

Inspired by the quadratic programming formulations of conventional control barrier functions in the continuous-time case~\cite{ames2016control}, we pose the controller synthesis problem as an optimization.

\textbf{CVaR Control Barrier Function Optimization:} At every time step $t$, given $x^t$, a set $\mathcal{S}$ as described in~\eqref{eq:safeset}, a confidence level $\beta \in (0,1)$, a parameter $\alpha \in (0,1)$, control upper bounds $\overline{u}$, lower bounds $\underline{u}$, and a legacy controller $u^t_{\text{legacy}}$, solve 
\begin{subequations} \label{eq:mainopt}
\begin{align}
    u^t_* = &\argmin_{u^t}~~ (u^t-u^t_{\text{legacy}})^T (u^t - u^t_{\text{legacy}}) \nonumber \\
            &\text{subject to}  \quad  \underline{u} \le u^t \le \overline{u}, \label{eq:controlbounds} \\
             & \quad \quad \quad   \quad   \mathrm{CVaR}_{\beta} \left( h(f(x^t,u^t,w))\right) \ge  \alpha h(x^{t}). \label{eq:conssafe}  
\end{align}
\end{subequations}
 Note that instantaneous controls $u^t$ are the only variables in the optimization. The cost function $(u^t-u^t_{\text{legacy}})^T (u^t - u^t_{\text{legacy}}) = \| u^t-u^t_{\text{legacy}}\|^2$ ensures that $u^t$ remains as close as possible to the legacy controller $u^t_{\text{legacy}}$ in the Euclidean $2$-norm; hence, it guarantees the \emph{minimal interference}.

For general nonlinear $h$, optimization problem~\eqref{eq:mainopt} is a nonlinear program in the decision variable $u^t$ (note that $\mathrm{CVaR}$ is a convex function in $h$ since it is a coherent risk measure). Indeed, this was the case for optimization problems designed for synthesizing discrete control barrier functions for discrete-time systems even without stochastic uncertainty~\cite{agrawal2017discrete}, as well.  MATLAB functions such as \texttt{fmincon} can be used to solve the nonlinear program. 

Next, we show that under some assumptions the search over CVaR-safe controls $u^t$ can be carried out by solving DCPs. For the remainder of this section, we restrict our attention to the case when system~\eqref{eq:dynamics} is a linear system. That is, $$f(x^t,u^t,w^t)=A(w^t)x^t+B(w^t)u^t+G(w^t),$$ where $A: \mathcal{W} \to \mathbb{R}^{n \times n}$, $B: \mathcal{W} \to \mathbb{R}^{n \times m}$ and $G: \mathcal{W} \to \mathbb{R}^{n}$.

For such systems, we assume the CVaR barrier function takes the form of a linear function
\begin{equation}\label{eq:rbflinear}
    h(x^t) = H x^t + l,
\end{equation}
where $H \in \mathbb{R}^{1\times n}$ and $l \in \mathbb{R}$. 
Then, the term~$\mathrm{CVaR}_{\beta} \left( h(f(x^t,u^t,w))\right)$ in constraints ~\eqref{eq:conssafe}  changes to 
\begin{equation} \label{eq:xcxvb}
\mathrm{CVaR}_{\beta}(H A(w)x^t+H B(w) u^t +HG(w) +l).
\end{equation}
Since $\mathrm{CVaR}_{\beta}$ is a convex function, the above term is a  convex function in $u^t$,\textit{ i.e.}, the control variable. 

Re-writing optimization problem~\eqref{eq:mainopt} for linear discrete time systems with stochastic uncertainty and CVaR barrier function~\eqref{eq:rbflinear} gives  the following optimization problem
\begin{subequations} \label{eq:mainoptlinear}
\begin{align}
    u^t_* = &\argmin_{u^t}~~ (u^t-u^t_{\text{legacy}})^T (u^t - u^t_{\text{legacy}}) \nonumber \\
            &\text{subject to}               \quad  \underline{u} \le u^t \le \overline{u}, \label{eq:controlbounds} \\
             & \quad \quad \quad   \quad \quad   \text{\eqref{eq:xcxvb}} \ge  \alpha h(x^{t}),  \label{eq:consreachL}
\end{align}
\end{subequations}

 Substituting the expression for CVaR~\eqref{eq:cvardual} in~\eqref{eq:xcxvb} for uncertainty $w$ with finite $|\mathcal{W}|$ yields  
\begin{multline}\label{cscseee}
\textstyle \inf_{\zeta \in \mathbb{R}} \bigg\{ \zeta +\frac{1}{\beta} \sum_{i=1}^{|\mathcal{W}|}  \big(H A(w_i)x^t \\
\textstyle +H B(w_i) u^t +HG(w_i) 
+l -\zeta\big)_{+}p(w_i)      \bigg\}, 
\end{multline}
which introduces the extra decision variable $\zeta \in \mathbb{R}$. 

Hence, \eqref{eq:mainoptlinear} can be rewritten in the standard DCP form 
\begin{subequations}\label{eq:DCPcvar}
\begin{align}
     u^t_* = & \argmin_{u^t,\zeta}~~ q_0(u^t)  \nonumber  \\
        &\text{subject to}  \quad q_1(u^t) \le 0~\text{and}~q_2(u^t) \le 0,  \\
        & \quad \quad \quad \quad \quad q_3 - \hat{q_4}(\zeta,u^t) \le 0, 
\end{align}
\end{subequations}
where $q_0(u^t) = (u^t-u^t_{\text{legacy}})^T (u^t - u^t_{\text{legacy}})$ is a convex (quadratic) function, $q_1(u^t)=\underline{u}-u^t$ is a convex (linear) function, $q_2(u^t)=u^t - \overline{u}$ is a convex (linear) function, and $q_3=\alpha H x^t +\alpha l$ is a convex (constant) function. The expression for $\hat{q_4}(\zeta,u^t)$ is given inside the braces in~\eqref{cscseee} which is a convex function in $u^t$ and $\zeta$ since  $\hat{q_4}(\zeta,u^t)$ is  convex in $\zeta$~\cite[Theorem 1]{rockafellar2000optimization} because the function $(\cdot)_+$ is increasing and convex~\cite[Lemma A.1.]{ott2010markov}.

 DCPs like \eqref{eq:DCPcvar} arise in  many applications, such as risk-averse MDPs~\cite{ahmadi2020constrained} and inverse covariance estimation in statistics~\cite{thai2014inverse}. In order to solve DCPs, we use a variant of the convex-concave procedure~\cite{shen2016disciplined}, wherein  the concave terms are replaced by a convex upper bound and solved. In fact, the disciplined convex-concave programming (DCCP)~\cite{shen2016disciplined} technique linearizes DCP problems into a (disciplined) convex program (carried out automatically via the DCCP  package~\cite{shen2016disciplined}). Then, the cone program can be solved readily by available solvers, such as 
YALMIP~\cite{yalmip}. 
 

We remark that solving~ \eqref{eq:DCPcvar} via the  DCCP method, finds the (local) saddle points to optimization problem~\eqref{eq:DCPcvar}. Nonetheless, every such local $u^t$ guarantees CVaR-safety.

\section{Case study: Bipedal Robot Locomotion}
Planning and controlling of bipedal walking are challenging problems, and there has been various related approaches \cite{grizzle2014models} in the literature. In this paper, we apply the approach in \cite{xiong2020ral} to approximate the bipedal walking dynamics via a discrete linear system subject to stochastic uncertainty. 



Let $\mathbf{x}_h = [c, p, v]^T$ denote the horizontal state, where $c$ is the horizontal position of the center of mass (COM) of the robot relative to the inertia frame, $p$ is the horizontal position of the COM relative to its stance foot, and  $v$ is the horizontal velocities of the COM. Then, the horizontal the step-to-step (S2S) dynamics~\cite{xiong2020ral} of the horizontal COM state is represented as 
\begin{equation}
\label{eq:robotS2S}
    \mathbf{x}_h^{t+1} = \mathcal{P}^h(x^t, \tau(t)),
\end{equation}
where $x$ is the full robot state and $\tau$ is the input torque. However, the S2S dynamics~\eqref{eq:robotS2S} cannot be obtained in analytical form due to the nonlinear and hybrid dynamics of the robot walking. 

Our previous work~\cite{xiong2020ral,xiong2021inReview} suggests that the S2S dynamics of the walking of the Hybrid-Linear Inverted Pendulum (H-LIP) \cite{xiong2021inReview} can be used to approximate the actual horizontal S2S dynamics of the walking of the robot in Eq. \eqref{eq:robotS2S}. The S2S dynamics of the H-LIP \cite{xiong2020ral} is: 
\begin{equation}
\label{eq:linearS2S}
    \mathbf{x}_{\text{H-LIP}}^{t+1} = A \mathbf{x}_{\text{H-LIP}}^{t} + B u_{\text{H-LIP}}^t
\end{equation}
where $\mathbf{x}_{\text{H-LIP}}^{t+1} = [c_{\text{H-LIP}}, p_{\text{H-LIP}}, v_{\text{H-LIP}}]^T$ is the discrete pre-impact state of the H-LIP, and $u_{\text{H-LIP}}^t$ is the step size. The expressions of $A, B$ can be found in \cite{xiong2020ral}. By approximation,  \eqref{eq:robotS2S} can be rewritten as
\begin{align} 
     \mathbf{x}_h^{t+1} &=  A \mathbf{x}_h^{t} + B u^t + w^t, \label{eq:approxlins2s}
\end{align}
where $w^t:= \mathcal{P}^h(x^t, \tau(t)) -  A \mathbf{x}_h^{t} -B u^t \in \mathcal{W}$ can be treated as a stochastic disturbance to the linear system in  \eqref{eq:linearS2S}. 


For application of 3D bipedal walking, the H-LIP model is applied in each plane of walking: the sagittal and the lateral planes. The H-LIP based planning provides the desired step sizes for the robot, which become the desired outputs for the low-level controller to track \cite{xiong2020ral}. 

 \begin{figure}[t]
      \centering
      \includegraphics[width = 0.75\columnwidth]{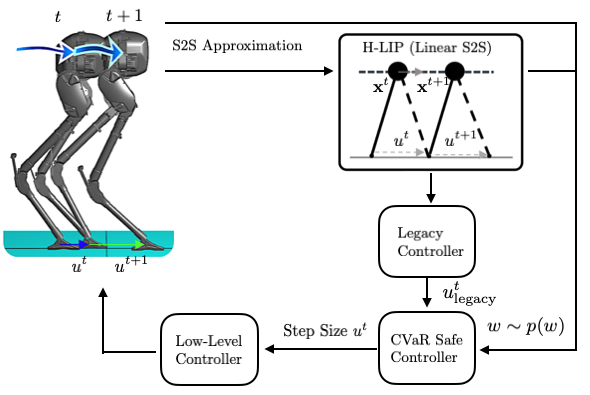}%
      \caption{A schematic diagram of our proposed risk-averse path planning method baaed on CVaR barrier functions for bipedal robots.}
      \label{fig:diagram}
\end{figure}

\subsection{Risk-Averse Bipedal Robot Path Planning}
We apply the CVaR barrier function based risk-averse planning presented in Section II to the 3D bipedal walking as described in Fig.~\ref{fig:diagram}. Model discrepancy $w$ is treated as stochastic uncertainty and as a risk factor that can lead to undesired behavior on the generated walking.


To circumvent this issue, we synthesize CVaR barrier functions based controllers to filter the H-LIP based stepping controller on the robot. The barrier functions are designed to represent the safe regions, where there are no obstacles. We are then interested in keeping the robot inside the safe (obstacle-free) regions during walking. 

The uncertainty $w$ is numerically calculated by several hours of simulations of different walking behaviors, which provided a polytopic set that bounds $w$. We took $|\mathcal{W}|$ random samples from the latter polytopic set.  Since $w$ is sparse in nature, we assumed a uniform distribution of $w$ inside $\mathcal{W}$,\textit{ i.e.}, $p(w)=1/|\mathcal{W}|$.  To design the risk-averse safe controllers, we then solve DCP~\eqref{eq:DCPcvar}, where $A$, $B$, and $G(w)=w$ are given by the approximated S2S dynamics~\eqref{eq:approxlins2s}. 

\subsection{Simulation Results}

We apply the proposed approach in high-fidelity simulation on the underactuated bipedal robot Cassie \cite{XiongSLIP}.  DCP~\eqref{eq:DCPcvar} is solved in YALMIP using MOSEK solver at each step. The optimization typically takes $100\sim 700$ steps under 10 seconds to solve on a laptop with the processor intel(R) Core(TM) i7-7700HQ@2.8GHz. The low-level controller on the robot is solved at 1kHz. The legacy controller used in our experiments is a model predictive controller. The simulation video of all the experiments can be found at \url{https://youtu.be/QNMW1zey3cI}.

\begin{figure}[t]
      \centering
      \includegraphics[width = .93\columnwidth]{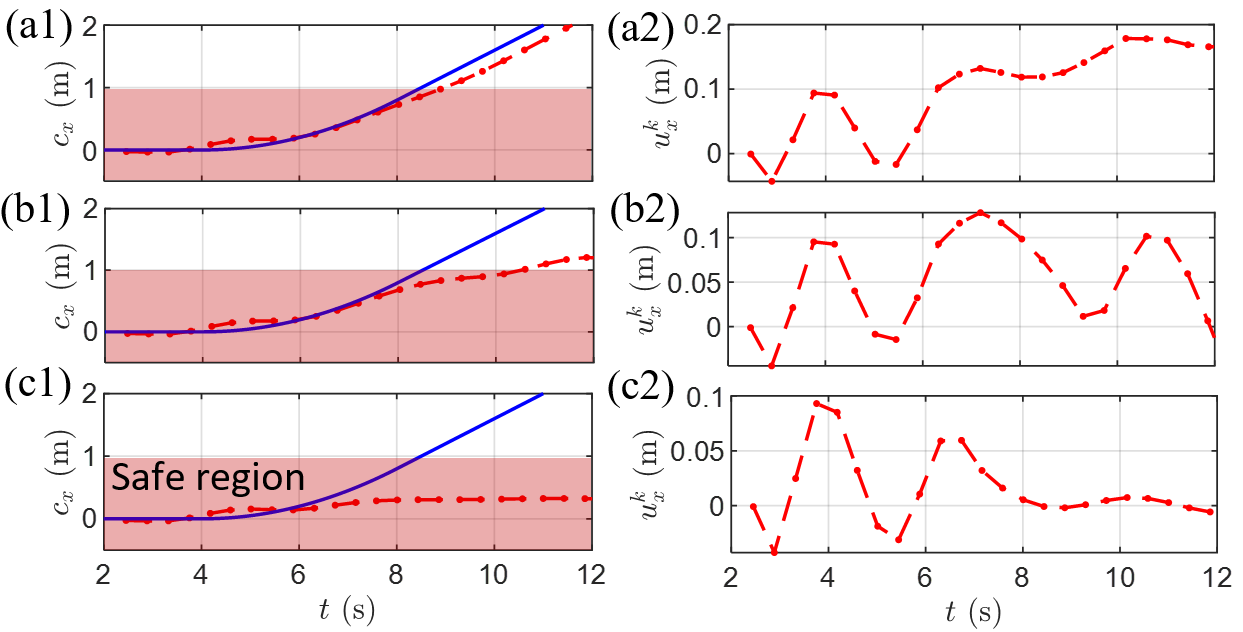}%
      \caption{\textit{Case 1}: Trajectories of the positions (blue is the desired trajectory) and step length of the robot in the sagittal plane for (a) walking without CVaR barrier function, where (a2) shows the legacy controller input, (b1) walking with risk neutral CVaR barrier function with $\beta=0.999$, where (b2) shows the output of the CVaR-safe controller with $\beta=0.999$, (c) walking with the risk-averse CVaR barrier function with $\beta=0.1$, where (c2) shows the output of the CVaR-safe controller with $\beta=0.1$. The red shaded area indicates the safe region for the robot.}
      \label{fig:result1}
\end{figure}

\textbf{Case 1:} We consider a scenario where the robot is following a straight path and an obstacle is placed in this path. The results are shown in Fig. 1 (a) and Fig. \ref{fig:result1}. The legacy controller is not aware of this wall, which results in collision that in practice would cause hardware failure. Then, we apply a CVaR barrier function to filter the output of the legacy controller. The safe set is defined as $$
    h(c_x) = p_x - c_x \geq 0,$$ 
where $p_x = 1$ is the position of the obstacle, $c_x$ denotes the position of the robot in the forward direction. We first apply the  
CVaR barrier function with $\beta=0.999$ (risk neutral) risk-averse controller. The result is shown in Fig. 1 (b): the robot walks and stop at the location of the obstacle. However, due the stochastic uncertainty $w$, the risk-neutral path planning violates the safety requirement. Lastly, we apply the CVaR barrier function with $\beta=0.1$ (risk-averse case), which generates the walking in Fig.~1~(c). The legacy controller directs the robot forward, but the CVaR-safe controller keeps the robot away from the obstacle. 

\begin{figure}[t]
      \centering
      \includegraphics[width = 0.99\columnwidth]{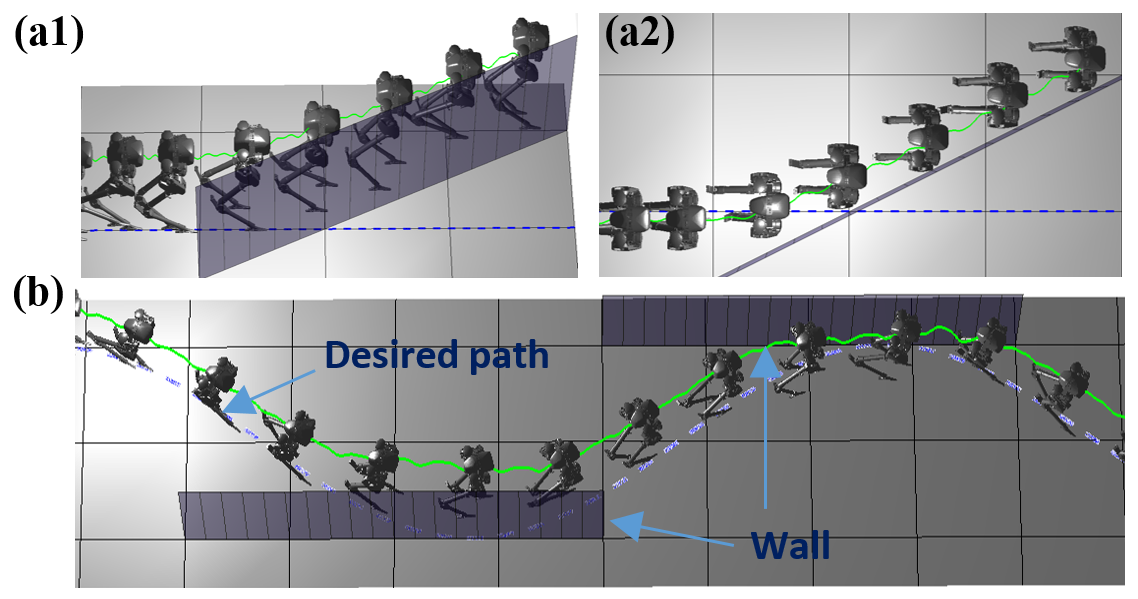}%
      \caption{The generated walking for \textit{Case 2} (a1, a2) and \textit{Case 3} (b). }
      \label{fig:result23}
\end{figure}

\textbf{Case 2:} In this scenario, we consider the robot following a  forward reference path. However, there is a wall at an angle, which does not completely prevent the robot from walking forward. The safe set is defined as $$
    h(c_x,c_y) = c_y  + k (c_x - p)  \geq 0, $$
where $k$ indicates the angle of the wall, $p$ indicates the location of the wall in forward direction, and $c_y$ is the position of the robot in the lateral plane. Here $k = -0.5$ and $p = 2$. Fig. \ref{fig:result23} (a) shows the generated walking behavior.  With the CVaR barrier function with $\beta=0.5$, the robot keeps a distance from the wall and maintains its original forward walking behavior in its sagittal plane, which is similar to the walking in Fig. \ref{fig:result1} (a1). As a result, the robot also walks laterally as well to assure CVaR-safety. 

\textbf{Case 3:} We consider a scenario with multiple barrier functions. The robot is supposed to follow a sinusoidal path. We add two walls on its way. The safe set is then defined as $\min(h_1,h_2) \ge 0$, where 
\begin{equation*}
    h_1(c_y) = c_y + p_1 \geq 0,~h_2(c_y) = -c_y + p_2\geq 0,
\end{equation*}
with $p_1 = 2$ and $p_2 = 0$. Fig. \ref{fig:result23} (b) illustrates the walking with the CVaR barrier function~$\beta=0.5$, where the robot successfully avoided the collision with the walls.

\section{Conclusion}

We proposed a method based on CVaR barrier functions to verify and enforce risk-averse safety for discrete-time stochastic systems. We proposed a computational method for synthesizing CVaR-safe controllers in the case of linear dynamics. The method was applied to enforce risk-averse safety of a bipedal robot. Future work will extend the CVaR barrier functions to other coherent risk measures, continuous-time systems, and  applications involving cooperative human-robot teams and imperfect sensor measurements~\cite{ahmadi2020risk}.


\bibliographystyle{IEEEtran}
\bibliography{references}

\end{document}